\documentclass[letterpaper, 10 pt, conference]{ieeeconf}  

\IEEEoverridecommandlockouts                              

\usepackage{balance}
\usepackage{hyperref}
\usepackage{graphicx}
\usepackage{amssymb}
\usepackage{amsmath}
\usepackage{amsfonts}
\usepackage{caption}
\usepackage{algorithm}
\usepackage{algorithmicx}
\usepackage[noend]{algpseudocode}
\usepackage{graphicx,caption}
\usepackage{amsmath,amssymb,stmaryrd,mathtools}
\usepackage{url}
\usepackage{subcaption}
\usepackage{booktabs}
\usepackage{multirow}
\usepackage{float}
\hypersetup{
    colorlinks=true,
    linkcolor=blue,
    filecolor=magenta,      
    urlcolor=cyan,
    hyperindex=True,
}
\title{\LARGE \bf
MANGA: Method Agnostic Neural-policy Generalization and Adaptation
}

\author{Homanga Bharadhwaj$^1$, Shoichiro Yamaguchi$^2$, and Shin-ichi Maeda$^2$
\thanks{$^1$ Work done while HB interned at Preferred Networks, Inc. Tokyo, Japan}
\thanks{$^1$ Homanga Bharadhwaj is with the Department of Computer Science, University of Toronto, Canada}%
\thanks{$^1$ Please address all correspondence to homanga@cs.toronto.edu}%
\thanks{$^2$ Shoichiro Yamaguchi and Shin-ichi Maeda are with Preferred Networks, Inc. Tokyo, Japan}
}

\begin{document}

\maketitle
\thispagestyle{empty}
\pagestyle{empty}

\begin{abstract}
In this paper we target the problem of transferring policies across multiple environments with different dynamics parameters and motor noise variations, by introducing a framework that decouples the processes of policy learning and system identification. Efficiently transferring learned policies to an unknown environment with changes in dynamics configurations in the presence of motor noise is very important for operating robots in the real world, and our work is a novel attempt in that direction.  We introduce MANGA: Method Agnostic Neural-policy Generalization and Adaptation, that trains dynamics conditioned policies and efficiently learns to estimate the dynamics parameters of the environment given off-policy state-transition rollouts in the environment. 
Our scheme is agnostic to the type of training method used - both reinforcement learning (RL) and imitation learning (IL) strategies can be used. We demonstrate the effectiveness of our approach by experimenting with four different MuJoCo agents and comparing against previously proposed transfer baselines.  
\end{abstract}

\section{Introduction}
One of the most well recognized goals of robotics research is to develop autonomous agents that can perform a wide variety of tasks in various complex environments. Recently numerous deep reinforcement learning (RL) and imitation learning (IL) based approaches have sought to achieve good performance in complex robotic tasks through minimal supervision. However, a major concern in experimenting with the real environment directly is safety, both of the robot and of the environment. Safety concerns and also the issue of reproducibility has drawn robotics research extensively to simulation environments.

An important benefit of simulators is that not only can we reset as many times as needed by varying the initial state and/or injecting stochastic noises such as observation noise and motor noise, but can also arbitrarily configure the environment. This enables us to change the dynamics parameters like mass, shape, size, and inertia of the agent,  friction between the agent and the environment, damping coefficients and gravitational acceleration. We leverage this to develop our approach such that a wide ensemble of simulation configurations can be used in training to achieve robustness to a new environment. We especially focus on adaptation to the unknown dynamics of the new environment.


Most previous approaches for transfer to different environments~\cite{splitnet,homanga,deepinverse,biped} have not explicitly taken advantage of the fact that we can dynamically sample a variety of environments in simulation, and some that have done so~\cite{dynamicsrand,domainrand,decoupling,strategyopt,noRML} do not attempt to learn an efficient off-policy scheme for inferring the dynamics of the environment. To remedy this, we adopt a two-fold approach, and claim the following contributions:
\begin{itemize}
    \item learn a good latent space by encoding observations through appropriate regularizations and explicitly concatenate to it an encoding of the vector of dynamics parameter configurations; condition the policy decoder on this latent representation
    \item develop a Bayesian meta-learning scheme to infer the dynamics parameter configuration of a given environment from a dataset of off-policy rollouts in that environment.
\end{itemize}

We demonstrate that by randomly sampling the parameters of the simulation environments, and adapting the policy to these varied configurations in training, we can achieve successful transfer at test time to a completely unseen dynamics configuration of the environment. An important point to note is that at test time, we do not have access to the ground truth system parameters. So, we develop a scheme to learn system parameters from random off-policy state transition data.

A desirable property of the transfer learning method is that it should be zero-shot in the sense that the transferred policy should not require any fine-tuning in the target environment so that safety of the real robot is not compromised, when applied to sim2real transfer. This can be indeed realized by our proposed approach.

Although we evaluate our model in simulation only and study transfer across different simulation environments, the approach can be extended to sim2real transfer settings as well, provided there is access to a real robot, and we can mimic the real dynamics well when we set appropriate dynamics parameters in the simulator. 


\section{Related Works}

Training robots directly in the real environment is unsafe, especially for domains like navigation/locomotion~\cite{safe3,safe2,safe1}, and hence training in simulation and deploying in the real world has become a common trend in robotics, under the theme of sim2real transfer~\cite{sim2real2,sim2real1}. An important first step to sim2real transfer is sim2sim transfer~\cite{splitnet,sim2sim1UPOSI}. Numerous recent works have tackled a similar problem as ours and studied transfer of policies across different simulation environments, across dynamics models, and from simulation to real environments. Universal Planning Network (UPN)~\cite{upn} trains for goal-directed tasks and in the process tries to capture `transferable representations' such that the trained encoders can be used for reward-shaping an RL algorithm for a similar albeit slightly complicated task. It is important to note that, complete on-policy training of the RL algorithm still needs to be performed in the new environment and the only `transfer' benefit provided by UPN is in reward shaping. Hence, a good transfer cannot be achieved zero-shot.

Learning a policy that is robust to dynamics change can be naively done by training a policy architecture in different domain randomized configurations. This has been done in the Domain Randomization~\cite{dynamicsrand,domainrand} approaches, the main drawback of which is that it learns an `average' policy that performs reasonably well in a wide range of test environments but is not `very good' for each environment. Motivated by this drawback, we do not aim to develop `robust' policies, but polices that can `adapt' to a given new test environment. A simple way to do this, as shown in~\cite{strategyopt} could be to maintain a repertoire of policies corresponding to different dynamics configurations (this ensemble is called \textit{the strategy}) and choose the best policy corresponding to the test environment, by running a few episodes, and considering the policy that yields the most rewards. However, since this method requires number of execution episodes in the test environment that is linearly proportional to the number of policies in \textit{the strategy}, the approach is not scalable.

In~\cite{openai}, the authors use LSTM~\cite{lstm} value and policy networks that implicitly learn the dynamics parameters of the environment during policy learning via dynamics randomization. However, learning a dynamics model together with policy learning renders less control over what is learned in the latent space and may lead to sensitive hyper-parameter optimization for achieving convergence. Hence, it is advisable to decouple the two procedures as advocated in~\cite{decoupling}. Another issue is the convergence of dynamics parameter estimation. Since LSTM assumes time-varying latent variables, and the observations change every time-step while the environmental dynamics remains fixed within an episode, trying to achieve convergence for both a good policy and an estimate of good dynamics may be difficult.


Meta-Learning~\cite{maml,bayesianmaml,finn2018probabilistic,prototypical} attempts to develop general models that can adapt to new tasks with a very few model updates. FastMAML~\cite{fastmaml} modifies MAML~\cite{maml} by separating the model parameters into general and task-specific parameters. Only the task-specific parameters need to be updated when a new `test' task is given. NoReward MAML~\cite{noRML} extends MAML~\cite{maml} to handle tasks defined by different dynamics configurations of the environment. The main difference from vanilla MAML is that the authors meta-learn the advantage function which is used to appropriately bias the Monte-Carlo sampling estimates of policy during learning. An important drawback of this approach is that by considering non-temporal state-action transition sequences (just static data of the form $(s_t,a_t,s_{t+1})$) important dynamics parameters like friction, gravity etc. cannot be appropriately modeled. 
Another drawback is that the method requires fine-tuning with some data samples in the test environment, and hence is not zero-shot.

Learning a domain-invariant latent space on which the policy is conditioned is another line of domain-adaptation based approaches for policy transfer. Zhang et al.~\cite{adversarial} adapt the encoder from sim to real by performing adversarial domain adaptation (ADA)~\cite{ada1,ada2} to match the latent space of encoding in sim and real, but require intermediate supervision in the form of position of robotic joints for the latent state while training the encoders. Bharadhwaj et al.~\cite{homanga} do this end-to-end without requiring intermediate supervision. However, both of these approaches suffer from the drawback of not being able to transfer effectively to different dynamics configurations as ADA cannot capture non-visual changes. Hence, they require fine-tuning in the real environment for aligning the dynamics modules, and so, are not zero-shot approaches.

Yu et al.~\cite{biped} adopt a two stage process for system identification and subsequently policy transfer is developed. The novelty of their method is in training a policy architecture conditioned on the roughly identified model parameters. However, a major concern of this approach is that on-policy state-transition data from the intermediately trained model is required in the target environment for system-parameter identification, which is not safe (since the model has not yet been fully trained).  Also, the method proposed in the paper can be used for transfer to a `fixed' target environment - when the target environment is altered i.e. the system dynamics parameters are altered, the entire method including `pre-SysID' needs to be re-trained. However our method, after being trained can be deployed on any test environment with unknown dynamics parameters and does not need to be re-trained when the test environments change.


\section{The Proposed Approach}
\label{sec:method}
The components of the proposed approach are described below:

\subsection{The basic model}
\begin{figure*}
    \centering
    \includegraphics[width=\textwidth]{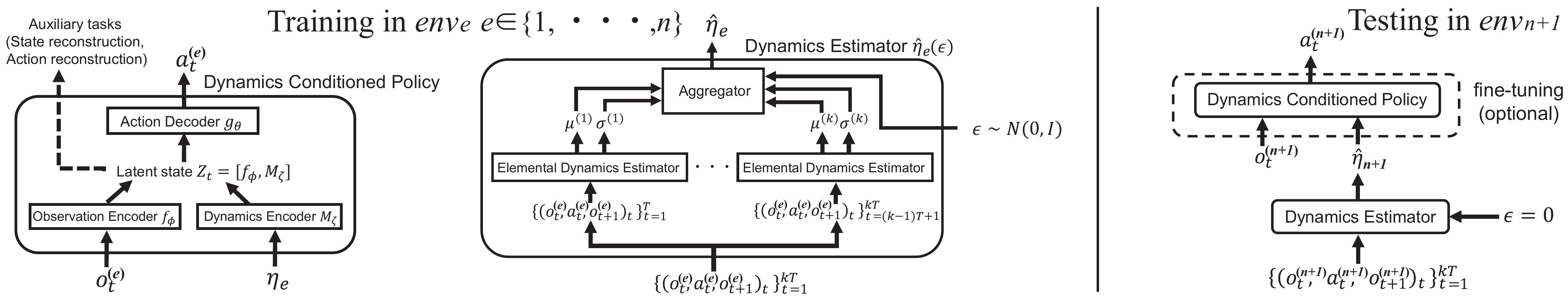}
    \caption{A schematic of the overall architecture of MANGA. The Elemental Dynamics Estimator and their aggregation is described in Sec.III D. }
    \label{fig:overall}
\end{figure*}

Our basic model consists of an encoder for observations and a policy (or `action') decoder. The Markov chain corresponding to the model is $X\longrightarrow Z\longrightarrow A$, where $X$ is the input state (which can either be fully observable or partially observable).  $A$ is the action space, a sample from which is what the model outputs. We consider $A$ to be a normal distribution whose mean and variance are predicted by the decoder from latent $Z$. Our training scheme is end-to-end and hence we do not need intermediate supervision for latent $Z$. In the subsequent sections, we denote the observation encoder by $f_\phi(\cdot)$ and the action decoder by $g_\theta(\cdot)$. Later, we also introduce the dynamics encoder, inverse dynamics model, and state (reconstruction) decoder respectively denoted by $M_\zeta(\cdot)$, $g_{inv}(\cdot)$, and $f_{rec}(\cdot)$. The \textit{Dynamics Conditioned Policy} module in Fig.~\ref{fig:overall} describes the basic architecture of MANGA. All the model components are realized by feed-forward neural networks.

\subsection{Dynamics conditioned policy (DCP)}
We condition our policy decoder both on the current observation frame in the environment and on an encoding of dynamics parameters of the agent and the environment. Many other previous papers~\cite{dynamicsrand,deepinverse} considered the raw dynamics parameters as input to the policy model (i.e. without encoding them separately from input observations), however, it is important to consider a separate encoding of the parameters so that they scale well and are in sync with the latent encoding of input observations. This is also important because the observations change in each time-step while the dynamics parameter vector, and thus the dynamics encoding, remains fixed within each episode of training.

Consider the process of training our model in a simulation environment $env_{e}$ and let the dynamics parameters of $env_{e}$ be denoted by a $d$-dim vector $\eta_e$. Now, we encode the ground-truth dynamics parameters through an encoder $M_{\zeta}$ and feed in the output $M_{\zeta}(\eta_e)$ to the bottleneck layer at time-step $t$ of our basic model. The bottleneck layer is the concatenation of $f_\phi(o_t^{(e)})$, where $o_t^{(e)}$ is the observation in $env_{e}$ at time $t$ and $M_{\zeta}(\eta_e)$ i.e. the vector $Z_t = [f_\phi(o_t^{(e)}),M_{\zeta}(\eta_e)]$. The policy decoder $g_\theta(\cdot)$ then takes as input the vector $Z_t$ and outputs the mean and covariance matrix for the action distribution. So,
\begin{align*}
    (\mathbf{m}_t,\mathbf{\Sigma}_t) = g_\theta([f_\phi(o_t^{(e)}),M_{\zeta}(\eta_e)]), \quad a_t^{(e)}\sim \mathcal{N}(\mathbf{m}_t,\mathbf{\Sigma}_t).
\end{align*}
Here $a_t^{(e)}$ is the output action of the model corresponding to the input observation $o_t^{(e)}$ and the dynamics vector $\eta_e$. 
The policy learned in $env_{e}$ is not likely to work well in the other environment $env_{e'}$, even if we provided the dynamics parameters $\eta_{e'}$, because we have not trained the policy to distinguish between the dependence on $\eta$ and $o_t$. To remedy this, we borrow the idea of dynamics randomization from Peng et al.~\cite{dynamicsrand}. 

\subsection{Training the DCP - Improving Generalization through Dynamics Randomization}
To implicitly learn the dependence of the policy on input observations $o_t$ and the dynamics $\eta$ of the environment, we train our model across different simulation environments $env_{e}$ $\forall e\in\{1,...,n\}$ by choosing random values for the dynamics parameters (within appropriate ranges) across the $n$ environments. At the start of each episode we sample a certain $\eta_e$ that defines an environment, choose a random initial pose (state) $o_1^{(e)}$ from the distribution of all states $\mathcal{I}^e$ and train the model corresponding to that environment and sample another $\eta_e$ randomly at the start of the next episode. This explicit conditioning over a wide ensemble of dynamics parameters enables transfer to unseen dynamics parameters. 

Our proposed method is agnostic to the type of training procedure, and both Reinforcement Learning (RL) and Imitation Learning (IL) approaches can be used. However, in the experiments we consider a specific RL algorithm for training, for the sake of consistency in comparison. 
The detailed training procedure for the dynamics conditioned policy is described through Algorithm~\ref{alg:training}.

\begin{algorithm}[h!]
	\caption{\textsc{Training Procedure of DCP} \label{alg:training}}
	\begin{algorithmic}[1]
		\Procedure{Train}{L-algo}
		\State Initialize params of $f_\phi,g_\theta,M_\zeta,f_{rec},g_{inv}$ 
		\State Generate $n$ randomized environments
		\For{epoch $i=1:K$}
		\For{each episode}
		\State Randomly choose environment $e\in\{1,..,n\}$
		\State Obtain dynamics parameters $\eta_e$
		\State Randomly sample the initial state $o_1^{e}\sim\mathcal{I}^e$
		\State Train $f_\phi,g_\theta,M_\zeta,f_{rec},g_{inv}$ using L-algo
		\EndFor
		\EndFor
		\EndProcedure
	\end{algorithmic}
\end{algorithm}

\subsection{Inferring the dynamics parameters at test time}

Since we do not have access to the system's dynamics parameters at test time, we propose a scheme to learn the system parameters from random off-policy state transition data. During training, we have access to the $n$ simulation environments and their corresponding dynamics parameter $\eta_e \in \mathbb{R}^d$ ($e=1, \cdots, n$). We consider a random policy that samples a random action in the range of allowed actions (or any pre-trained `safe' policy) and allow it to run for a few episodes in each environment. We collect state transition data in tuples of the form (state, action, next state) i.e., $\{(o_t^{(e)},a_t^{(e)},o_{t+1}^{(e)})\}$. Let $\mathcal{F}(\cdot)$ be a forward dynamics model of the simulator such that $ \hat{o}_{t+1}^{(e)} = \mathcal{F}(o_{t}^{(e)},a_{t}^{(e)}; \eta_e)$ where the true next state is given as
\begin{equation}
     o_{t+1}^{(e)} = \mathcal{F}(o_{t}^{(e)},a_{t}^{(e)}; \eta_e) + R_t, \label{eq:likelihood}
\end{equation}
when we have the true value of system parameters $\eta_e$. $R_t$ is a noise term, and for the sake of analytic simplicity, we assume $R_t$ is a Gaussian with zero mean and variance $v^2$. The above defines the likelihood model for state-transition $Q(o_{t+1}^{(e)}|o_{t}^{(e)},a_{t}^{(e)}, \eta_e)$ and our aim is to estimate $\eta_e$ through its posterior $p(\eta_e|D_e)$ where $D_e =\{(o_t^{(e)},a_t^{(e)},o_{t+1}^{(e)})_t\}_{t=1}^{N_e} $. 

Although some previous approaches~\cite{noRML} try to estimate system dynamics parameters from uncorrelated, stand-alone state-transition tuples, we postulate that to correctly estimate dynamics parameters, we must consider correlated state-transition data within episodes. We divide the horizon length of the episodes in each $env_{e}$ into $k$ chunks of length $T$ each and estimate $\eta_e$ for each chunk $i\in\{1,..,k\}$ in the form of Gaussian distribution with mean $\mu^{(i)}$ and variance $\sigma^{(i)}$ by using an elemental dynamics estimator.
If we denote the observation sequence and action sequence within the chunk $i$ in $env_{e}$ as $y_i^{(e)}$ and $x_i^{(e)}$, this amounts to the estimate of the following posterior: 
\begin{align*}
    p(\eta_e | x_{i}^{(e)}, y_{i}^{(e)}) \propto  p(y_{i}^{(e)}|x_{i}^{(e)},\eta_e)p(\eta_e)
\end{align*}

The length of chunk, $T$, should be large, but not be too large. 
To aggregate the $k$ estimates of $\eta_e$s, we exploit the relationship between the posterior of $\eta_e$ conditioned on a single pair of datapoints, $p(\eta_e | x_{i}^{(e)}, y_{i}^{(e)})$, and the posterior of $\eta_e$ conditioned on the entire dataset $D_e$, $p(\eta_e | D_e)$. We note that:
\begin{align*}
    p(\eta_e | D_e) &\propto  \{\prod_{i=1}^k p(y_{i}^{(e)}|x_{i}^{(e)},\eta_e)\}p(\eta_e)\\
    &\propto  \left\{ \prod_{i=1}^{k}p(\eta_e | x_{i}^{(e)}, y_{i}^{(e)}) \right\}/ p(\eta_e)^{k-1}
\end{align*}

Because we assume $p(\eta_e | x_{i}^{(e)}, y_{i}^{(e)})$ and $p(\eta_e)$ to both be independent Gaussian distributions, $p(\eta_e | D_e)$ can be obtained as a Gaussian distribution after some elementary computations,
\begin{align}
p(\eta_e | D_e) = \prod_{j=1}^d N(\eta_{e,j}| \mu_j(D_e), \sigma_j(D_e)), \label{eq:Gauss_pos}
\end{align}
where 
\begin{align*}
    \mu_{j} = \sigma^{2}_{j} \left( \sum_{i=1}^{k} \frac{\mu_{j}^{(i)}( x_{i}^{(e)}, y_{i}^{(e)})}{\sigma_{j}^{(i)}( x_{i}^{(e)}, y_{i}^{(e)})^2} -(k-1)\frac{f_{0,j}}{g_{0,j}^2}\right)
\end{align*}
\begin{align*}
    \sigma^{-2}_{j} = \left(\sum_{i=1}^{k}\sigma_{j}^{(i)}(x_{i}^{(e)}, y_{i}^{(e)})^{-2}\right) - (k-1)g_{0,j}^{-2}
\end{align*}

Here the subscript $j$ denotes the $j^{th}$ element of the vector. 
The posterior $p(\eta_e | D_e)$ is parameterized by $\theta$ as $p_{\theta}(\eta_e | D_e)$  through the parameterization of the functions $\mu_{j}^{(i)}(\cdot)$ and $\sigma_{j}^{(i)}(\cdot)$ which are realized by deep neural networks. Also $\theta$ includes scalars $f_{0,j}$ and $g_{0,j}$ ($j=1,\cdots,d$). The parameters $\theta$ are optimized so as to approximate the true posterior well.

A popular way of approximating the posterior is to minimize the KL divergence between the true posterior $p_{\rm{true}}(\eta_e| D_e)$ and its approximation $p_{\theta}(\eta_e| D_e)$, i.e.,
\begin{align*}
    \min_{\theta} KL[p_{\theta}(\eta_e| D_e) || p_{\rm{true}}(\eta_e| D_e)]
\end{align*}
This posterior approximation problem for each environment $env_{e}$ can be solved without explicitly evaluating the $p_{\rm{true}}(\eta_e| D_e)$ when we consider the following evidence lower bound optimization~\cite{vae}. 

\footnotesize
\begin{align*}
 &\max_{\theta} 
  \sum_{n=1}^{N_e} \log  p_{\rm{true}}(y_{n}|x_{n},\eta_e)
 - KL[p_{\theta}(\eta_e| D_e) | p_{\rm{true}}(\eta_e| D_e)]  \\
=&\max_{\theta} \int p_\theta(\eta_e | D_e) \sum_{n=1}^{N_e}\log p_{\rm{true}}(y_{n}|x_{n},\eta_e)  d\eta_e \\&- KL[p_\theta(\eta_e | D_e)|p_\theta(\eta_e)]
\end{align*}
\normalsize
Here we can use the re-parameterization trick~\cite{vae} that replaces the expectation with respect to $p_{\theta}(\eta_e| D_e)$ with the expectation with a standard Gaussian variable $\epsilon \sim N(\epsilon|{\rm{0}},{\rm{I}})$ by interpreting the Gaussian posterior $p_{\theta}(\eta_e| D_e) $ to be a result of element-wise variable transformation 
$\hat{\eta}_{e,j} (\epsilon_j) =  \mu_j +\sigma_j \epsilon_j$ with
$\epsilon_j \sim N(\epsilon_j|0,1)$.
\footnotesize

\begin{align*}
    \max_{\theta}\frac{1}{2v^2}\mathop{\mathbb{E}}_{\epsilon\sim N(\epsilon|{\rm{0}},{\rm{I}})}\left[\sum_{i=0}^{k-1}\sum_{t=1}^{T}[o_{iT+t+1}^{(e)} - \mathcal{F}(o_{iT+t}^{(e)},a_{iT+t}^{(e)}; \hat{\eta}_e(\epsilon)]^2 \right]\\ 
    + \frac{1}{2}\sum_{j=1}^d \left( \frac{(\mu_j-f_{0,j})^2 + \sigma_j^2}{g_{0,j}^2} + \log\frac{g_{0,j}^2}{\sigma_j^2}\right),
\end{align*}
\normalsize

It is important to consider a chunk of temporal sequences instead of standalone tuples $(o_t^{(e)},a_t^{(e)},o_{t+1}^{(e)})$ (unlike~\cite{noRML}) so as to effectively realize the posterior of complex dynamics parameters like friction, gravity etc. In general, the posterior of the dynamics parameter can take a complex multi-modal distribution, but it approaches a Gaussian when the no. of samples in the temporal chunk increases and the statistical model is `regular'  according to the central limit theorem \cite{watanabe2009algebraic}. The $k$ different estimates of $\eta_e$ from temporal chunks of length $T$ each in each episode of the rollouts are the `Elemental Dynamics Estimator' in Fig.~\ref{fig:overall}. Their `Aggregator' is described by the optimization problem above.



Given state-transition data, we can use the trained model to infer the dynamics parameter vector of the test environment. It is important to note that collecting data in the test environment for this system parameter identification is inexpensive because we only need {\it{off-policy}} data, which can be collected by simply running a random policy or a different pre-trained `safe' policy.


\subsection{Test Time inference}
At test time we are given a simulation environment $env_{n+1}$ with dynamics parameters $\eta_{n+1}$ which are unknown. Let $f_\phi,g_\theta,M_\zeta,f_{rec},g_{inv}$ denote our trained model components that have been adapted through training in $n$ different dynamics configurations. Our aim now is to transfer the policy that has been learned in training, without any fine-tuning in the test environment i.e. we are not allowed to train again in $env_{n+1}$. We can do this by using the learned $\hat{\eta}_{n+1}$ in lieu of ground-truth $\eta_{n+1}$ and running forward inference through the trained model $f_\phi,g_\theta,M_\zeta$. This scheme is demonstrated in Algorithm~\ref{alg:testing}. Although our approach learns a very good zero-shot initialization in the test environment (Section IV), we show comparisons with other models that require on-policy fine-tuning in the test environment in Section IV. Fine-tuning corresponds to updating the parameters of the policy architecture while executing in the test environment. For MANGA, to achieve good zero-shot initialization, the only execution in the test environment needed is running a random policy (or some external trained policy) to collect state transition data for feeding into the trained dynamics estimation module.
\begin{algorithm}[h!]
	\caption{\textsc{Test Time Inference} \label{alg:testing}}
	\begin{algorithmic}[1]
		\Procedure{Test}{TrainedParams}
		\State Initialize $f_\phi,g_\theta,M_\zeta,f_{rec},g_{inv}$  with TrainedParams \par and denote the test environment $e=n+1$
	   \State Observe off-policy state transition data $D_{n+1}$
		\State Estimate dynamics parameters $\hat{\eta}_{n+1}(D_{n+1})$ 
		\State {Execute the policy  from given initial state $o_1^{(n+1)}$ \par with the model $f_\phi,g_\theta,M_\zeta$}
		\EndProcedure
	\end{algorithmic}
\end{algorithm}
\subsection{Adapting to variations in motor noise}
In this section we discuss a scheme to make our model robust to motor noise, which is an important consideration for real robotic tasks~\cite{deepinverse}. 
We interpret the addition of motor noise as a form of domain randomization,
 and consider that in reality we have some specific state dependent deviation. The implication of motor noise is same as adding disturbance to the output action of our policy model. 
In order to infer a model for the disturbance $\epsilon_t$, we assume it to be a function of the current state $o_t^{(e)}$ weighted by an environment dependent parameter $\omega_e$. Hence, $\epsilon_t=\omega_e\Phi_\tau(o_t^{(e)})$, where $\Phi$ is a non-linear mapping, specifically a feed-forward neural network whose parameter $\tau$ have been randomly assigned and fixed (similar random networks have been used for exploration and uncertainty estimation in RL~\cite{randomnetwork}).
When $\omega_e$ is randomly set with a large enough output dimension $d$ of $\Phi_\tau(\cdot)$ during the training of the policy, the training scheme under this motor noise is similar to a form of domain randomization. 
However, we actively identify the perturbation caused by this in environment $e$ through the estimation of $\omega_e$.

Let the original predicted action at time-step $t$ be $\hat{a}_t^{(e)}$. The action that is fed to the simulator, because of the motor noise now becomes  $\hat{a}_t^{(e)}+\epsilon_t= \hat{a}_t^{(e)} + K\omega_e\Phi_\tau(o_t^{(e)})$. Since $\omega_e$ is an environment dependent parameter just like $\eta_e$, we estimate the concatenated vector $\eta_e'=(\eta_e,\omega_e)$ through the scheme described in Section III D for estimating $\eta_e$. Here $K$ is a scalar multiplier to the noise.




\subsection{State reconstruction and ignoring nuisance correlates}
Since we are training policies adaptable to variations in the environment, we need to ensure that our agent's policy does not unfairly make correlations with state changes like changes in brightness, direction of light, location of shadow etc. that occur not as a result of the policy. Previous works like~\cite{dynamicsrand} do not consider this issue, however we argue that it is important for the very reason that we consider randomized environments. To tackle this and avoid learning nuisance correlates, we enforce an inverse dynamics model based regularization, which was previously used in the Intrinsic Curiosity Module of~\cite{curiosity}. Let $z_t=[f_\phi(o_t), M_\zeta(\eta)]$ be the latent state at timestep $t$, and $g_{inv}$ be the inverse dynamics model such that the predicted action $\hat{a}_t=g_{inv}(z_{t+1},z_{t})$. The loss function is: $ \mathcal{L}_{inv}^{(t)} = (g_{inv}(z_{t+1},z_{t}) - a_t)^2$.

In addition to the regularization via an inverse dynamics model, we also enforce input state reconstruction from the learned latent representation. This is important because we do not want our policy to get conditioned on a latent state which can never be reached from the observation space. Thus we aim to learn a reconstruction $\hat{o}_t=f_{rec}(f_\phi(o_t))$ such that $(\hat{o}_t - o_t)^2$ is minimized. The loss function is: $    \mathcal{L}_{rec}^{(t)} = (f_{rec}(f_\phi(o_t))-o_t)^2$




\section{Experiments}
Through a series of experiments, we demonstrate the necessity of the different components of the proposed MANGA approach (ablation study) and compare against some external baselines for adaptation to different dynamics at test time. We also experimented to see how adaptive is MANGA to the change in the range of the dynamics parameter variations and how it adapts to motor noise variations at test time.

\begin{figure*}[!htbp]
\centering
\includegraphics[width=0.85\linewidth]{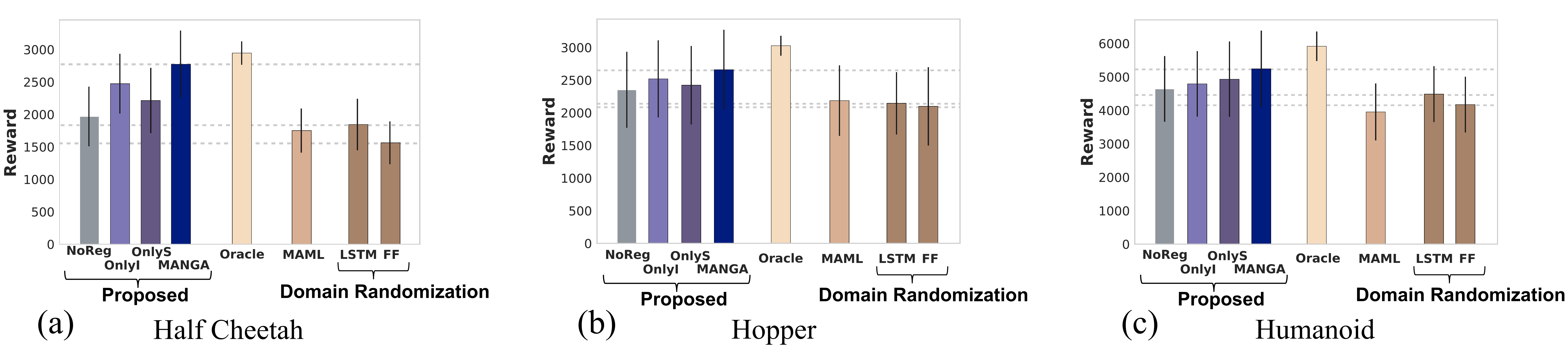} 
\caption{Ablation study and comparison. All models are trained for 200,000 episodes and fine-tuned for 100 episodes in the test environment for effective comparison with MAML.  Results are averaged over 1,000 episodes of execution in the unseen test environment with dynamics parameters in the range of $\pm5\%$ of base values. NoReg is vanilla MANGA without any additional regularization. OnlyI is MANGA sans the state reconstruction regularization. OnlyS is MANGA sans the inverse dynamics regularization. LSTM is the DR baseline trained with LSTM policy architecture to implicitly estimate $\eta$ during policy learning~\cite{openai}. FF is the DR baseline corresponding to MANGA without any regularization and with no $\eta$ estimation.}
\label{fig:ablation}
\end{figure*}
\begin{figure*}[ht]
\centering
 \includegraphics[width=0.9\textwidth]{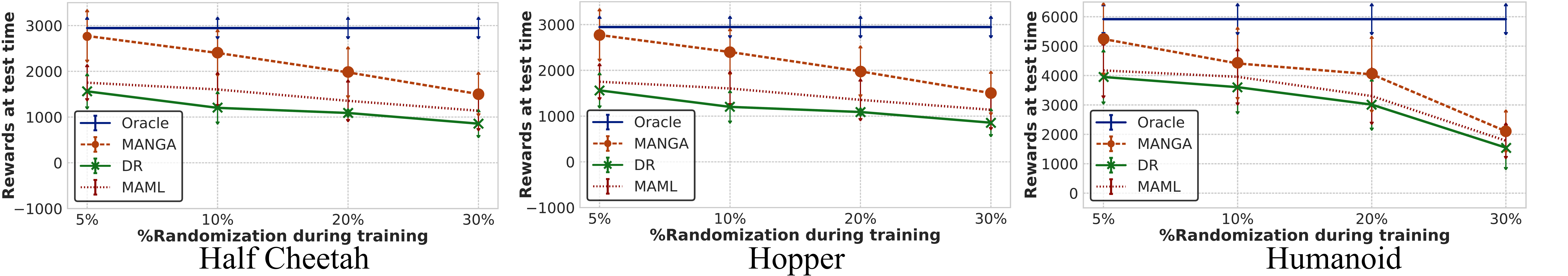}   
\caption{Training with different ranges of system parameters for 200,000 episodes. The evaluation is on a randomly chosen previously unseen test environment within the same respective range. For effective comparison with MAML, all the models are updated in the test environment for the same number of episodes (100) as MAML. Higher reward is better.}
\label{fig:different_randomization}
\end{figure*}

\begin{figure*}[!htbp]
\centering
\includegraphics[width=0.95\textwidth]{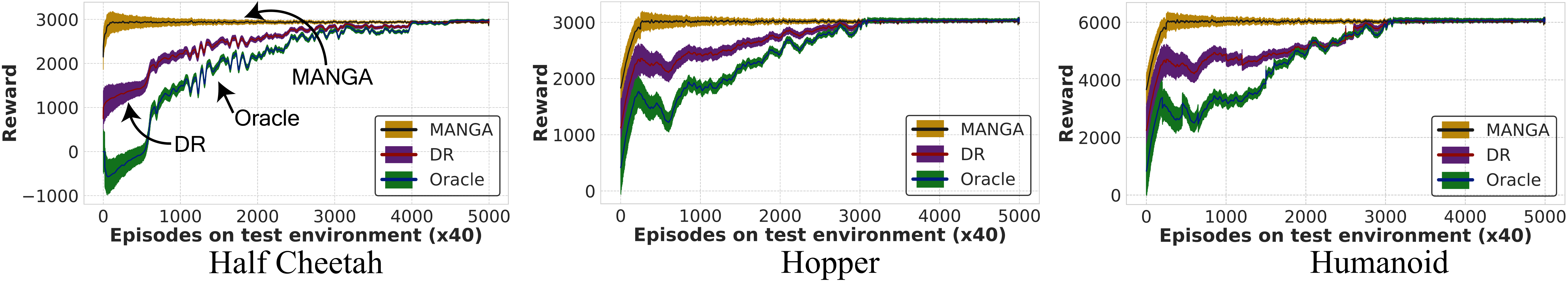}
\caption{Plot of training with rollouts in the unseen test environment with dynamics parameters in the range of $\pm5\%$ of base values. MANGA is the model that has been trained by the proposed  approach for 200,000 episodes. 200 episodes of the random policy are used to estimate the dynamics parameters of the test environment. DR is the Domain Randomization baseline (FF) corresponding to our model. Oracle is the version of our model that is directly trained in the test environment and has access to the true dynamics parameters of the test environment.}
\label{fig:testrollouts}
\end{figure*}

\subsection{MuJoCo Environments (OpenAI Gym)}
We consider three different MuJoCo environments~\cite{mujoco} of varying complexity - Humanoid-v2, HalfCheetah-v2, and Hopper-v2, where the task in each environment is to move the agent as fast as possible without toppling over~\cite{openaigym}. For consistency in comparison with external baselines, we use the default reward setting for each environment as specified in~\cite{openaigym} and alter the following dynamics variables for evaluation: mass ($\mathcal{M}$) and inertia  ($\mathcal{I}$) of the agent, gravitational acceleration ($\mathcal{G}$), friction coefficient between the agent and the environment ($\mathcal{F}$), stiffness coefficient of joints ($\mathcal{S}$), and damping coefficient ($\mathcal{D}$).

Each dynamics variable $\mathcal{M},\mathcal{I},\mathcal{G},\mathcal{F}, \mathcal{S}, \mathcal{D}$ for MuJoCo is in general a vector (for example the $\mathcal{M}$ vector consists of the mass of different parts of the Half-Cheetah body) of different dimensions. We consider $\eta$ to be the linearized concatenation of all these variables. Let $\eta^{i}$ corresponds the $i^{th}$ dynamics variable in $\eta$, whose base value is say ${\bar{\eta}}^{i}$. 

During training, we randomize $\eta$ such that each component $\eta^{i}$ gets perturbed in the range $[{\bar{\eta}}^{i}-x{\bar{\eta}}^{i}, {\bar{\eta}}^{i}+x{\bar{\eta}}^{i}]$. Here $x$ denotes $\%$ randomization and we perform experiments with randomly chosen $\eta$ in the respective range specified by $x$. 


\subsection{Training details}
Although our proposed approach is method-agnostic, and L-algo in Algorithm~\ref{alg:training} can be any RL or IL algorithm, for our specific implementation we used the RL algorithm Proximal-Policy Optimization (PPO) algorithm~\cite{ppo}. We used SGD~\cite{sgd} optimizer for optimization and the Pytorch~\cite{pytorch} library in Python for the implementation.  For training the dynamics estimator in Fig.~\ref{fig:overall}, we found that choosing a temporal chunk $T$ of length $50$ timesteps (for each elemental dynamics estimator) performed well. All the functions described in Fig.~\ref{fig:overall} and Section~\ref{sec:method} are realized by feed-forward neural networks. Other details including the baselines are described in the subsequent sections.

\subsection{Ablation study}
We postulate that the auxiliary modules, namely the inverse dynamics model and the state-reconstruction decoder are needed to learn a good latent space $Z$ for effective transfer. 
MANGA refers to the proposed approach with all components present.
For reference, we compare MANGA with an Oracle. Oracle refers to an untrained agent of the same architecture as MANGA, that has access to ground-truth system parameters, and is trained from scratch directly in the test environment.
In Fig.~\ref{fig:ablation}, we show results by selectively ablating different components of the proposed model when the dynamics parameters are perturbed in the range of $\pm5\%$ of base values. There is clearly a drop in performance on the test environment when we remove either or both the auxiliary modules. Interestingly, removing the inverse dynamics model causes a very sharp decrease in performance across all the three MuJoCo domains. Hence, it is clear that ignoring nuisance correlates between states and actions is important for quick and effective transfer.

\subsection{Comparisons with existing methods in literature}
We compare the performance of MANGA with existing approaches in a new environment whose dynamics parameters are perturbed in the range of $\pm5\%$ of base values. Note that $\pm5\%$ perturbation is large enough to cause significant performance drop when a version of MANGA was trained in the base environment only and then tested in the randomized test environment without any adaptation ($1010.12\pm247.33$,  $956.15\pm361.83$, $2041.58\pm694.75$ respectively for Half-Cheetah, Hopper, and Humanoid) . The results are in Fig.~\ref{fig:ablation}.

We consider two external baselines, namely Domain Randomization (DR) and meta-learning. For DR, we followed the implementation of the state-of-the art dynamics randomization paper~\cite{dynamicsrand} with two variants LSTM and FF. LSTM is the variant that uses an LSTM policy and value architecture while implicitly identifying the system dynamics parameters during policy learning~\cite{openai}. FF is with the same policy architecture (FF) as our MANGA model and without any system parameter identification. Since the LSTM variant is computationally very expensive and takes a long time to train, we perform only one type of comparison against it. For meta-learning, we implemented No-Reward MAML~\cite{noRML}, which performs significantly better than vanilla MAML~\cite{maml} for the scenario of transfer to different environments dynamics.  To ensure fair comparison, all models were trained for the same number of episodes, by executing for the same number of maximum time-steps per episode and the same optimizer was used for the rest of all experiments. 
\subsection{Analysis with different randomization ranges}
The extent to which we need to randomize the dynamics parameters during training depends on how different the test environment is likely to be with respect to the default setting. We experimented with different test environments in the range of $5\%, 10\%, 20\%, 30\%$ maximum variation of dynamics parameters from the default value. The range of randomized environment during training was also the same ($5\%, 10\%, 20\%, 30\%$ respectively) in each case.

As evident from Fig.~\ref{fig:different_randomization}, the performance of all the compared models decrease when the range of parameter variations is increased. However, the drop in performance of MANGA is the least compared to the other methods. We attribute this favourable behavior, primarily to the fact that we have separated the processes of system parameter identification and policy learning with regularization. Hence, the latent space learned for conditioning the policy is not potentially negatively affected by the training of the system parameter identification module.

\subsection{Quick Adaptation: Rollouts in the test environment}
Although most approaches for policy transfer~\cite{noRML,strategyopt,decoupling} need rollouts in the test environments for reasonably good transfer, our proposed approach adapts a good policy zeros-shot by estimating dynamics parameters based on  the observation of random off-policy state transition data, as shown by the reward at episode $0$ of the plots in Fig.~\ref{fig:testrollouts}. 
Furthermore, we observe that if allowed to update model parameters in the test environment (i.e. fine-tuning), MANGA quickly converges and achieves reward equivalent to the Oracle only within a few hundred  episodes.  

\subsection{Evaluation of performance in the presence of Motor Noise}
We consider two variants of MANGA here: MANGA-Noise and MANGA-NoNoise. MANGA-Noise has been trained by considering random values of $\omega_e$ corresponding to each randomized environment, and a fixed $\tau$ (i.e. the weights of the random network are fixed) during training. We learn a model for estimating the value of $\omega_e$ along with $\eta_e$ as described in Sec III D and F. At test time we consider two situations of motor noise - \textit{known noise} and \textit{unknown noise}. \textit{Known noise} corresponds to the case when at test time, the value of $\tau$ is same as that during training, while \textit{unknown noise} corresponds to the case when the value of $\tau$ at test-time is different from that during training. 

\begin{figure}[!htbp]
    \centering
    \includegraphics[width=9cm]{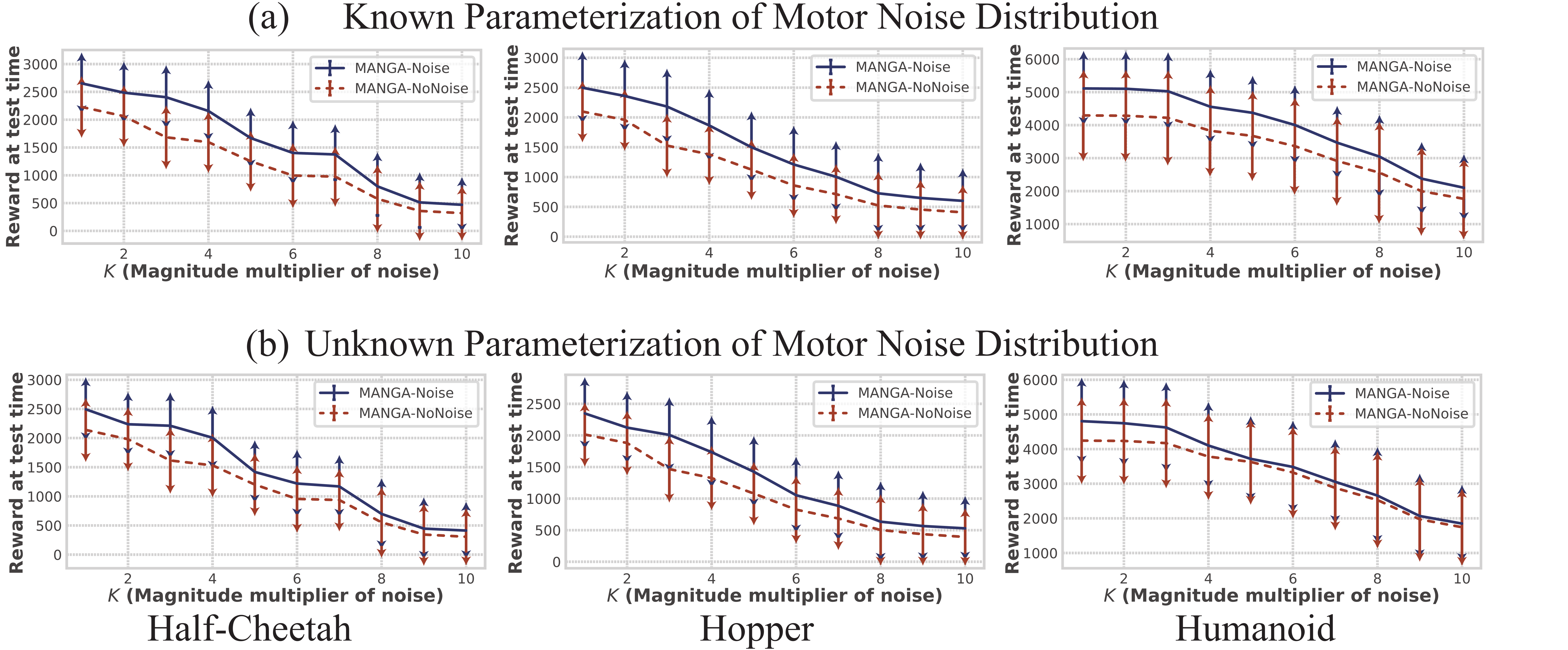}  
    \caption{Evaluation of variants of MANGA in the presence of motor noise in the unseen test environment with dynamics parameters in the range of $\pm5\%$ of base values, after 200,000 episodes of training. MANGA-Noise corresponds to the case when motor noise is present in training and $\omega_e$ is inferred as described in Section III F. MANGA-NONoise corresponds to the case when motor noise is present during training but encoding of $\omega_e$ is not input to the latent $Z$ and $\omega_e$ for the test env is not inferred. $K$ denotes the magnitude of the noise multiplier. The top row corresponds to the scenario of known noise ($\tau$ same as in training). The bottom row corresponds to the scenario of unknown noise ($\tau$ is randomly chosen to be different from training).}
    \label{fig:noise}
\end{figure}

It is evident from Fig.~\ref{fig:noise} that MANGA-Noise effectively estimates the weight vectors $\omega_e$ and achieves much higher reward than MANGA-NoNoise in the presence of noise in the test environment. This suggests the effectiveness of the noise estimation technique described in Section III F. 

\section{Conclusion}
In this paper, we introduced a general framework for policy transfer that decouples the processes of policy learning and system identification, is agnostic to the algorithm used for training it and can quickly adapt to an environment at test time with variations in dynamics and motor noise. We compared the proposed approach with existing algorithms for policy transfer and demonstrated its efficacy with respect to robustness to the range of dynamics variations, variation in motor noise, quick adaptation to a test environment and learning of a transferable latent space for policy conditioning. 
\section{Acknowledgement}
We would like to acknowledge the support of Crissman Loomis-san, Takashi Abe-san,  Masanori Koyama-san, Yasuhiro Fujita-san and many other colleagues at Preferred Networks Tokyo who helped shaped this work through the amazing research discussions during Homanga Bharadhwaj's internship. We also thank Florian Shkurti (University of Toronto) for his valuable feedback on the draft and help in editing it. 

\bibliographystyle{IEEEtran}
\bibliography{references}

\end{document}